\pdfoutput=1
\documentclass[11pt]{article}
\usepackage[final]{acl}
\usepackage{times}
\usepackage{latexsym}
\usepackage[T1]{fontenc}
\usepackage[utf8]{inputenc}
\usepackage{microtype}
\usepackage{inconsolata}
\usepackage[most]{tcolorbox}
\usepackage{graphicx}
\usepackage{caption}
\usepackage{float}
\usepackage{titlesec}
\usepackage{array}
\usepackage{algorithm}
\usepackage{algpseudocode}
\usepackage{amsmath}
\usepackage{multicol}
\usepackage{amssymb}
\usepackage{amsfonts}
\usepackage{booktabs}
\usepackage{multirow} 
\usepackage{arydshln}
\usepackage{ragged2e}
\usepackage{pgfplots}
\usepackage[export]{adjustbox}
\usepackage{subcaption}
\usepackage{xcolor}
\usepackage{bm}
\usepackage{fontawesome5}
\pgfplotsset{compat=1.18}

\title{AGIC: Attention-Guided Image Captioning to Improve Caption Relevance}

\author{
    L. D. M. S. Sai Teja\textsuperscript{1} \enspace 
    Ashok Urlana\textsuperscript{2} \enspace 
    \textbf{Pruthwik Mishra}\textsuperscript{3} \\
    \textsuperscript{1}NIT Silchar \enspace
    \textsuperscript{2}TCS Research, Hyderabad \enspace
    \textsuperscript{3}SVNIT Surat, India \\
    \texttt{lekkalad\_ug\_22@cse.nits.ac.in, ashok.urlana@tcs.com}, \\ [1ex] 
    \href{https://image-caption-relevance.github.io/AGIC/}{\faGithub\ \texttt{https://image-caption-relevance.github.io/AGIC/}}
}

\begin{document}
\maketitle
\vspace*{-1cm}

\begin{abstract}

Despite significant progress in image captioning, generating accurate and descriptive captions remains a long-standing challenge. In this study, we propose Attention-Guided Image Captioning (AGIC), which amplifies salient visual regions directly in the feature space to guide caption generation. We further introduce a hybrid decoding strategy that combines deterministic and probabilistic sampling to balance fluency and diversity. To evaluate AGIC, we conduct extensive experiments on the Flickr8k and Flickr30k datasets. The results\footnote{\url{https://github.com/saitejalekkala33/AGIC-code}} show that AGIC matches or surpasses several state-of-the-art models while achieving faster inference. Moreover, AGIC demonstrates strong performance across multiple evaluation metrics, offering a scalable and interpretable solution for image captioning.

\end{abstract}

\section{Introduction}
Image captioning, a prominent task in computer vision, aims to generate a visually grounded description of an image \cite{cornia2020meshed, cornia2019show, chen2015microsoft}. While significant performance improvements have been achieved, current methods frequently generate generic captions, limiting their utility in capturing nuanced visual details \cite{al2025ensemble, ma2023towards, guo2020recurrent, lu2018neural}. The core challenge in producing relevant and descriptive image captions stems from the inherent difficulty in comprehensively capturing every visual aspect within an image.

\begin{figure}[t]
    \centering
    \resizebox{1.0\linewidth}{!}{%
    \begin{minipage}[c]{0.5\textwidth}
        \centering
        \includegraphics[width=0.75\linewidth]{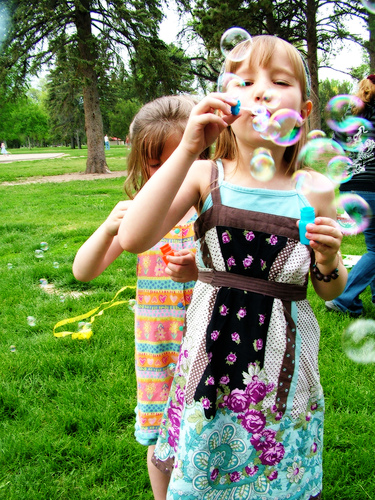}
    \end{minipage}%
    \begin{minipage}[c]{0.32\textwidth}
        \RaggedRight
            \textbf{\textit{\textcolor{red}{LlaVA:}}} Two little girls are playing with bubbles in the park.\\
            \textbf{\textit{\textcolor{red}{Qwen:}}} Two young friends share a joyful moment, creating a world of bubbles in the park.\\
            \textbf{\textit{\textcolor{red}{Fuyu:}}} Two little girls blowing bubbles in the park.\\
            \textbf{\textit{\textcolor{cyan}{BRNN:}}} Two girls playing in the park.\\
            \textbf{\textit{\textcolor{cyan}{LSTNet:}}} A young girl is blowing bubbles, holding them in her hands.\\
            \textcolor{violet}{\textit{\textbf{$\bm{R}^2\bm{M}$:}}} Two young girls playing with bubbles in grass.\\
            \textbf{\textit{\textcolor{blue}{AGIC}:}} \textcolor{magenta}{Two young girls wearing floral dress blowing bubbles in a park covered with grass.}
    \end{minipage}
    }
    \caption{Comparison of various image caption generation models. \textcolor{red}{red:} zero-shot, \textcolor{cyan}{cyan:} supervised, \textcolor{violet}{violet:} unsupervised approaches and \textcolor{blue}{blue:} our approach.} 
    \label{fig:page-1}
    \vspace{-7mm}
\end{figure}

To address limitations in caption relevancy and descriptiveness, existing approaches are broadly categorized into three main types: supervised, unsupervised, and semi-supervised methods. Supervised methods rely on large-scale, manually annotated image-caption pairs for training \cite{xu2015show, cornia2019show, anderson2018bottom}. While they often yield highly accurate captions, they are resource-intensive, requiring significant time and costly “gold-standard” data. In contrast, unsupervised methods utilize unpaired image sets and independent text corpora to learn visual concept detection and cross-modal alignment without explicit image-caption pairings \cite{feng2019unsupervised, guo2020recurrent, laina2019towards}. However, these models typically generate less precise or coherent captions due to the absence of direct supervision. Finally, semi-supervised methods \cite{liang2024generative} attempt to bridge this gap by generating pseudo-labels for unlabeled data, which are then used to guide the learning process. Although less data-hungry, these methods can suffer from error propagation if the pseudo-labels are inaccurate.

To overcome these challenges, in this study, we propose a novel attention-guided training-free approach by amplifying the pretrained model's attention weights to capture the relevant regions of the image along with a hybrid decoding approach. The proposed approach offers a compelling alternative by significantly reducing reliance on costly paired datasets, making them particularly attractive for scenarios where extensive ground-truth data is scarce. As shown in Figure~\ref{fig:page-1}, the zero-shot and unsupervised-based approaches tend to generate generic captions, whereas the supervised methods are unable to capture most of the relevant objects present in the image. In contrast, the proposed AGIC approach covers all the relevant objects present in the image and provides grammatically fluent and descriptive caption.

Our key contributions are:
1) A novel attention-guided image caption approach designed to generate relevant and descriptive captions. 2) A novel hybrid decoding strategy for enhanced caption generation. 3) Extensive experiments conducted on two popular image caption datasets using several Vision-Language Models (VLMs), along with rigorous ablation studies.

\section{Related Work}
\label{rel_work}
Image captioning is a challenging task, and several efforts have been made to solve this problem.  A family of attention-based approaches \cite{anderson2018bottom, huang2019attention, cornia2020meshed, sharma2018conceptual} have been incorporated for the object detection by capturing attributes, and other characteristics in images. Visual attention-based approaches focus on salient regions in an image using an object relation transformer that captures spatial dependencies between the detected objects \cite{herdade2019image}. In the same line, \citet{anderson2018bottom} propose bottom-up and top-down attention to combine object-level visual features with textual words. \citet{huang2019attention} extends this using Attention-on-Attention (AoA) with an additional gating mechanism to determine the relevance of the final attention results and the incoming queries. 

More recently, to infuse relevancy in image captioning, \citet{honda-etal-2021-removing} tackle spurious word-level alignments in pseudo-captioning by implementing a gating mechanism to filter out irrelevant visual words. \citet{shi-etal-2021-enhancing} develops a descriptive image captioning model using Natural Language Inference (NLI) between all pairs of reference captions. 
To better capture semantic information, \citet{shi-etal-2020-improving} constructs visual relationship graphs guided by captions. They also develop a multitask and weakly multi-instance learning framework for accurate explicit object and predicate detection. 
Another notable contribution by \citet{pan2020x} involves the modification of the conventional attention block into X-Linear attention networks to learn second-order feature interactions using bilinear pooling. 

Although supervised approaches have achieved state-of-the-art performance, they rely on large paired datasets, whereas unsupervised efforts \cite{feng2019unsupervised, gu2019unpaired, honda-etal-2021-removing} suffer from semantic drift and coherence. In contrast, we propose a training-free approach by amplifying the pretrained model's attention weights to capture the relevant regions of the image and generate descriptive captions.  

\section{Model Description}
\label{model_desc}
We propose a framework to improve image caption relevance using a contextual relevance amplification mechanism, implemented through an attention-guided process. Our approach is inspired by \citet{liu2025attention}, who used attention patterns and self-reflection to detect hallucinations in large language models without relying on labeled data.

\begin{table*}[h]
\centering
\scriptsize
\begin{tabular}{llcccccccc}
\toprule
\textbf{Dataset} & \textbf{Approach} & \textbf{B1} & \textbf{B2} & \textbf{B3} & \textbf{B4} & \textbf{R-L} & \textbf{METEOR} & \textbf{CIDEr} & \textbf{SPICE} \\
\midrule
\multirow{10}{*}{\textbf{Flickr8k}} 
    & BLIP2-opt-2.7b \cite{li2023blip} & 0.391 & 0.255 & 0.163 & 0.102 & 0.325 & 0.259 & 0.077 & 0.041 \\
    & LLaVA-1.5B-7B \cite{liu2023visual} & 0.440 & 0.293 & 0.193 & 0.128 & 0.357 & 0.049 & 0.198 & 0.080 \\
    & Qwen2.5-VL-7B-Instruct \cite{bai2025qwen2}& 0.441 & 0.276 & 0.170 & 0.107 & 0.311 & 0.034 & 0.280 & 0.072 \\
    & Fuyu-8B \cite{fuyu-8b} & 0.630 & 0.448 & 0.302 & 0.201 & 0.147 & 0.441 & 0.414 & 0.12\\
    & BRNN \cite{karpathy2015deep} & 0.563 & 0.362 & 0.219 & 0.131 & 0.350 & 0.171 & 0.512 & 0.112 \\
    & $R^2M$ \cite{guo2020recurrent} & 0.496 & 0.302 & 0.177 & 0.081 & 0.320 & 0.132 & 0.284 & 0.030 \\
    & LSTNet \cite{ma2023towards} & 0.669 & 0.448 & 0.304 & 0.241 & 0.417 & 0.210 & 0.623 & 0.137 \\
    & Ensemble \cite{al2025ensemble} & \textbf{0.728} & 0.495 & 0.323 & 0.208 & 0.432 & 0.235 & 0.604 & 0.164 \\
    \cmidrule{2-10} 
    & Ours (AGIC + LLaVA-1.5B-7B) & 0.651 & 0.478 & 0.342 & 0.239 & \textbf{0.442} & 0.242 & 0.697 & \textbf{0.213} \\
    & \textbf{Ours (\textit{AGIC + BLIP2-opt-2.7b})} & 0.665 & \textbf{0.499} & \textbf{0.355} & \textbf{0.251} & 0.248 & \textbf{0.445} & \textbf{0.734} & 0.195 \\
    \midrule
\multirow{10}{*}{\textbf{Flickr30k}}  
    & BLIP2-opt-2.7b \cite{li2023blip} & 0.406 & 0.260 & 0.164 & 0.103 & 0.306 & 0.245 & 0.083 & 0.083 \\
    & LLaVA-1.5B-7B \cite{liu2023visual} & 0.471 & 0.316 & 0.213 & 0.145 & 0.348 & 0.070 & 0.234 & 0.096 \\
    & Qwen2.5-VL-7B-Instruct \cite{bai2025qwen2} & 0.454 & 0.277 & 0.166 & 0.100 & 0.285 & 0.055 & 0.254 & 0.088 \\
    & Fuyu-8B \cite{fuyu-8b} & 0.613 & 0.427 & 0.288 & 0.193 & 0.387 & 0.166 & 0.396 & 0.134 \\
    & BRNN \cite{karpathy2015deep} & 0.559 & 0.340 & 0.194 & 0.113 & 0.317 & 0.153 & 0.499 & 0.109 \\
    & $R^2M$ \cite{guo2020recurrent} & 0.531 & 0.328 & 0.192 & 0.117 & 0.359 & 0.137 & 0.181 & 0.083 \\
    & LSTNet \cite{ma2023towards} & 0.654 & 0.493 & 0.350 & 0.224 & 0.410 & 0.197 & \textbf{0.633} & 0.310 \\
    & Ensemble \cite{al2025ensemble} & \textbf{0.798} & \textbf{0.561} & \textbf{0.387} & \textbf{0.269} & \textbf{0.443} & 0.213 & 0.565 & \textbf{0.387} \\
    \cmidrule{2-10}
    & Ours AGIC + LLaVA-1.5B-7B & 0.629 & 0.449 & 0.315 & 0.216 & 0.382 & 0.215 & 0.553 & 0.153 \\
    & \textbf{Ours (\textit{AGIC + BLIP2-opt-2.7b})} & 0.650 & 0.469 & 0.336 & 0.235 & 0.232 & \textbf{0.399} & 0.601 & 0.164 \\
\bottomrule
\end{tabular}
\caption{Image captioning results for various models; B1 to B4 refer to BLEU scores and R-L: ROUGE-L score.}
\label{tab:combined_sota_comparison}
\vspace{-5mm}
\end{table*}

\subsection{Attention Weights Extraction}
\label{first_pass}
In the AGIC framework, input images are passed to a pre-trained vision transformer model to obtain the corresponding attention weights of all image features. These attentive weights help to identify the most relevant regions of an image. 

Let $X^{l-1} \in \mathbb{R}^{N \times d}$ represent the patch embeddings at layer $l-1$, where $N$ is the number of image patches and $d$ is the embedding dimension. The attention matrix $A^{l,h} \in \mathbb{R}^{N \times N}$ at layer $l$ and head $h$ is computed as:

\vspace{-2mm}

{\footnotesize
\begin{equation}
\label{attn_at_lh}
A^{l,h} = \mathrm{softmax} \left( \frac{(X^{l-1} W_Q^{l,h})(X^{l-1} W_K^{l,h})^T}{\sqrt{d_h}} \right)
\end{equation}
}

Where $W_Q^{l,h}$ and $W_K^{l,h}$ are the query and key projection matrices at layer $l$, head $h$, and $d_h$ is the head dimension. Then, to aggregate attention weights across all heads for a specific layer $l$, the attention weights received by the visual patch $i$ in layer $l$ are given as follows.

\begin{equation}
\label{mead_head_attn}
a_{i}^{l} = \frac{1}{H} \sum_{h=1}^{H} A^{l,h}_{i}
\end{equation}

Where $H$ is the total number of attention heads.

\subsection{Image Amplification}
\label{inbetween_passes}
To make all the relevant features in the image more prominent, we perform the image amplification step using the attention weights. For amplification, we multiply attention weights of all the image features with the original image representation with an amplification factor \textit{k} as shown in the equation~\ref{amplify_tensor}.
\begin{equation}
\label{amplify_tensor}
{I}_{a}(i,j) = I_{o}(i,j) \cdot \left(1 + k \cdot a(i,j) \right)
\end{equation}

Where ${I}{a}(i,j)$ and $I{o}(i,j)$ denote the amplified and original values at the spatial location $(i,j)$, respectively. The term $a(i,j)$ represents the attention weights obtained during the first pass, and $k$ is a hyperparameter that controls the strength of amplification. We obtain the attention-guided image representation ${I}_{a}$ and pass it into the image captioning model to generate attention-guided captions.

\subsection{Caption Generation}
\label{sec_pass}
 To enhance both diversity and fluency in the caption generation for the amplified image, we adopt a \textit{hybrid decoding strategy} that performs stochastic beam search by combining beam search with Top-$k$, Top-$p$ (nucleus) sampling, and temperature scaling. At each decoding step, candidate tokens are sampled from a probability distribution obtained by temperature-scaled softmax, followed by Top-$k$ and Top-$p$ filtering. Sampling is performed independently within each beam, and the final caption is selected from the best completed beam.

Formally, for the token logit distribution $\mathbf{z}_t$ at decoding step $t$, we sample the next token $x_t$ as:
    \vspace{-3mm}

\begin{equation}
    x_t \sim \text{Top-}p\left(\text{Top-}k\left(\text{Softmax}\left(\frac{\mathbf{z}_t}{T}\right)\right)\right)
\end{equation}

where $T$ is the temperature and the sampling is carried out within each of the $B$ beams, which represents the number of parallel decoding paths to enhance contextual relevance in the generated captions. This enables our pipeline to focus on generating contextually relevant regions that lead to more detailed and meaningful image captions.

\section{Experiments and Results Analysis}
\subsection{Setup}
 In this study, we utilized two popular image caption datasets, Flickr8k \cite{hodosh2013framing} and Flickr30k \cite{you2016image}, to conduct the image captioning experiments. To perform the evaluation of the models, we utilize the BLEU ($n= 1, 2, 3, 4$) \cite{papineni-etal-2002-bleu}, ROUGE-L \cite{lin-2004-rouge}, METEOR \cite{banerjee-lavie-2005-meteor}, Consensus-based Image Description Evaluation (\textit{CIDEr}) \cite{vedantam2015cider}, and Semantic Propositional Image Caption Evaluation (\textit{SPICE}) \cite{anderson2016spice} metrics. 

\subsection{Baselines}
\label{comparision}
We compare our approach with comprehensive collection of baselines categorized as follows: (1) \textit{zero-shot prompting-based} approaches by utilizing the popular vision language models such as BLIP2-opt-2.7b \cite{li2023blip}, LLaVA-1.5B-7B \cite{liu2023visual}, Qwen2.5-VL-7B-Instruct \cite{bai2025qwen2} and Fuyu-8B \cite{fuyu-8b}, (2) \textit{unsupervised method} $R^2M$~\cite{guo2020recurrent}, (3) \textit{supervised methods} BRNN~\cite{karpathy2015deep}, LSTNet \cite{ma2023towards}, and Ensemble \cite{al2025ensemble}. We compare all these methods with our AGIC, and the corresponding results are detailed in Table~\ref{tab:combined_sota_comparison}. More details on the experimental setup can be found in Appendix~\ref{sec:experimental_setup}.
\subsection{Results Analysis}
In our AGIC approach, we utilize the BLIP2-OPT-2.7B \cite{li2023blip} and  LLaVA-1.5B-7B  \cite{liu2023visual} models for the caption generation, and as shown in Table~\ref{tab:combined_sota_comparison}, the proposed AGIC outperforms other baseline methods across all metrics on the Flickr8k dataset. On the Flickr30k dataset, AGIC achieves performance comparable to state-of-the-art methods. Furthermore, as shown in Appendix~\ref{sec:inference_time}, Figure~\ref{fig:inference_time}, our approach achieves significantly lower inference time compared to other state-of-the-art methods, offering a more cost-effective and time-efficient solution.
\section{Ablation study}
\textbf{Varying the layers:} To validate the contribution of the attention patterns of various layers, we generate the captions with the first, middle, last, max, and mean attention layers. We find that the attention scores obtained from the mean of the layers outperform their counterparts as detailed in Appendix~\ref{sec:ablation_more} Table~\ref{tab:diff_layers}.

\noindent \textbf{Varying the amplification factor:} As depicted in Table~\ref{tab:combined_ablation_studies}, we vary the amplification factor (k) and for both datasets, k=1 resulted in better performance. 
\begin{table}
\centering\footnotesize
\setlength{\tabcolsep}{0.6ex}
\begin{tabular}{lllccccccc}
\toprule
 &  & \textbf{Config.} & \textbf{BLEU} & \textbf{MTR} & \textbf{R-L} & \textbf{CIDEr} & \textbf{SPICE}\\
\midrule
\multirow{10}{*}{\rotatebox{90}{\textbf{Decoding strategy}}}
& \multirow{5}{*}{\rotatebox{90}{Flickr8k}}
& BLIP2-opt & 0.08 & 0.13 & 0.28 & 0.19 & 0.10\\
& & $+$Top-k & 0.08 & 0.13 & 0.29 & 0.30 & 0.13\\
& & $+$Top-p & 0.11 & 0.17 & 0.32 & 0.38 & 0.15\\
& & $+$Beam Search & 0.24 & 0.23 & 0.43 & 0.68 & 0.18\\
& & \textbf{All} & \textbf{0.25} & \textbf{0.24} & \textbf{0.44} & \textbf{0.73} & \textbf{0.19}\\
\cmidrule{2-8}
& \multirow{5}{*}{\rotatebox{90}{Flickr30k}}
& BLIP2-opt & 0.18 & 0.21 & 0.36 & 0.43 & 0.13\\
& & $+$Top-k & 0.07 & 0.14 & 0.23 & 0.24 & 0.10\\
& & $+$Top-p & 0.10 & 0.15 & 0.27 & 0.29 & 0.11\\
& & $+$Beam Search & 0.22 & 0.21 & 0.37 & 0.54 & 0.15\\
& & \textbf{All} & \textbf{0.23} & \textbf{0.23} & \textbf{0.39} & \textbf{0.60} & \textbf{0.16}\\
\midrule
\multirow{8}{*}{\rotatebox{90}{\textbf{Amplification factor}}}
& \multirow{4}{*}{\rotatebox{90}{Flickr8k}}
& AGIC (k=1) & \textbf{0.25} & \textbf{0.24} & \textbf{0.44} & \textbf{0.73} & \textbf{0.19}\\
& & AGIC (k=3) & 0.23 & 0.23 & 0.43 & 0.66 & 0.17\\
& & AGIC (k=5) & 0.23 & 0.23 & 0.42 & 0.56 & 0.16\\
& & AGIC (k=10)& 0.22 & 0.22 & 0.42 & 0.64 & 0.15\\
\cmidrule{2-8}
& \multirow{4}{*}{\rotatebox{90}{Flickr30k}}
& AGIC (k=1) & \textbf{0.23} & \textbf{0.23} & \textbf{0.39} & \textbf{0.60} & \textbf{0.16}\\
& & AGIC (k=3) & 0.21 & 0.20 & 0.37 & 0.52 & 0.14\\
& & AGIC (k=5) & 0.21 & 0.20 & 0.37 & 0.55 & 0.13\\
& & AGIC (k=10)& 0.21 & 0.20 & 0.37 & 0.51 & 0.12\\
\bottomrule
\end{tabular}
\caption{\textbf{Top:} Performance variation across decoding strategies. `All' represents the combined strategy (Beam Search + Top-$k$ + Top-$p$). \textbf{Bottom:} Performance comparison with varying amplification factors.}
\label{tab:combined_ablation_studies}
\vspace{-6mm}
\end{table}

\noindent \textbf{Varying the decoding strategy: } We experiment with different decoding strategies such as Top-k sampling, Top-p (nucleus) sampling, and beam search, combining all of these resulted in better performance compared to individual decoding strategies as shown in Table~\ref{tab:combined_ablation_studies}.

\begin{table}[t]
\centering\footnotesize
\setlength{\tabcolsep}{0.8ex}
\vspace{-3mm}
\begin{tabular}{llll} \\ \toprule
                & \begin{tabular}[c]{@{}l@{}}\textbf{Correctness} \\ \textbf{(Corr)} \end{tabular} & \begin{tabular}[c]{@{}l@{}} \textbf{Completeness} \\ \textbf{(Comp)} \end{tabular} & \begin{tabular}[c]{@{}l@{}} \textbf{Relevance} \\ \textbf{(Rele)} \end{tabular} \\
Mean            & 4.04                        & 3.64                         & 4.26                      \\
\midrule
                & \textbf{Corr–Comp}          & \textbf{Comp–Rele}           & \textbf{Rele–Corr}                 \\ 
Correlation (r) & 0.72                        & 0.81                         & 0.79                      \\ \bottomrule
\end{tabular}
\caption{Human evaluation scores and Correlations.}
\label{tab:human_eval}
\vspace{-6mm}
\end{table}

\section{Qualitative Assessment}
To assess the quality of image captions generated by our AGIC approach, we perform human evaluations and error analysis on 50 random samples. 
\subsection{Human Evaluation and Error Analysis}
\label{sec:human_error}
Our human evaluation focuses on \textit{Correctness, Completeness, Relevancy} metrics, and each is scored on a scale of 1 to 5.
Table~\ref{tab:human_eval} presents the average human evaluation scores and inter-metric correlations. Captions are rated highly overall, with Relevancy achieving the highest average (4.26). Correlation analysis shows strong alignment among metrics, particularly between Completeness and Relevancy (r $\approx$ 0.81), suggesting consistency in human assessment of caption quality.
Captions with the highest human scores often demonstrated precise object identification, contextually appropriate action descriptions, and inclusion of secondary and background scene details. Additional details on the human evaluation can be found in Appendix~\ref{sec:human_appendix}.

We conduct a detailed error analysis by manually examining four aspects: hallucination, omission, irrelevance, and ambiguity. While AGIC generally improves caption relevance, our analysis reveals that it occasionally omits salient objects present in the image and tends to hallucinate on Flickr30k data samples. These observations highlight areas for further refinement. Additional examples and discussion are provided in Appendix~\ref{sec:ablation_error}.

\begin{table}[htb]
\centering\footnotesize
\setlength{\tabcolsep}{0.6ex}
\begin{tabular}{@{}lllll@{}}
\toprule
          & Hallucination & Omission & Irrelevance & Ambiguity \\ \midrule
\textbf{Flickr8k}  & 7             & 12       & 3           & 5         \\
\textbf{Flickr30k} & 11            & 14       & 3          & 4         \\ \bottomrule
\end{tabular}
\caption{Error analysis of AGIC model.}
\label{tab:error_cater}
\vspace{-7mm}
\end{table}

\section{Conclusion}
\label{conclusion}
We propose \textit{AGIC}, an unsupervised image captioning framework that uses attention weights from different layers of a Vision Transformer to amplify contextually relevant regions for enhanced caption generation. \textit{AGIC} integrates an attention-guided amplification process with a hybrid decoding strategy to balance grounding and diversity with relevance to the image. Experimental results on standard benchmarks show that AGIC performs better than supervised and unsupervised baselines on several metrics, providing the most relevant captions. 

\section{Limitations}
While AGIC exhibits impressive performance when compared to several state-of-the-art models, it has the following limitations. 1) Due to reliance on attention allocation patterns, the applicability of the approach is restricted to only open-source models. 2) AGIC is highly sensitive to the amplification factor; this suggests that even slight over-amplification can be detrimental, diluting focus rather than enhancing it.

\bibliography{custom}

\appendix

\section{AGIC Pipeline Overview with an Illustrative Example}

\begin{figure*}[t]
    \centering
    \includegraphics[width=0.8\linewidth]{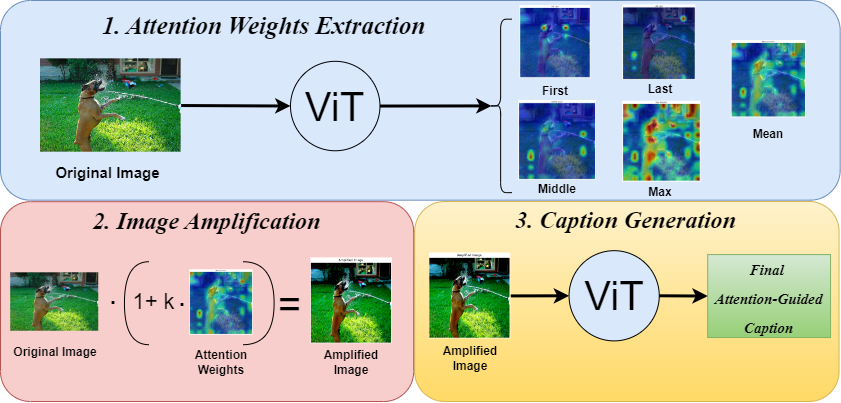}
    \caption{Attention guided image captioning (AGIC) pipeline.}
    \label{fig:agic_workflow}
\end{figure*}

The AGIC framework operates in three key stages:

\begin{itemize}
    \item \textbf{Attention Weights Extraction:} In the first stage, the model identifies and extracts attention weights corresponding to semantically relevant objects within the image. As illustrated in Figure~\ref{fig:agic_workflow}, the mean attention is concentrated around the object of interest, in this case, the `dog', indicating strong model focus.

    \item \textbf{Image Amplification:} Using the extracted attention weights, the image is selectively amplified by applying an amplification factor to regions associated with high attention. This step enhances salient features, making them more prominent for the captioning model.

    \item \textbf{Caption Generation:} Finally, the amplified image is input into a Vision Transformer (ViT) to produce a detailed and contextually accurate caption that aligns closely with the amplified regions of interest.
\end{itemize}

\section{Details on Inference time comparison}
\label{sec:inference_time}
\begin{figure}[htb]
    \centering
    \includegraphics[width=1\linewidth]{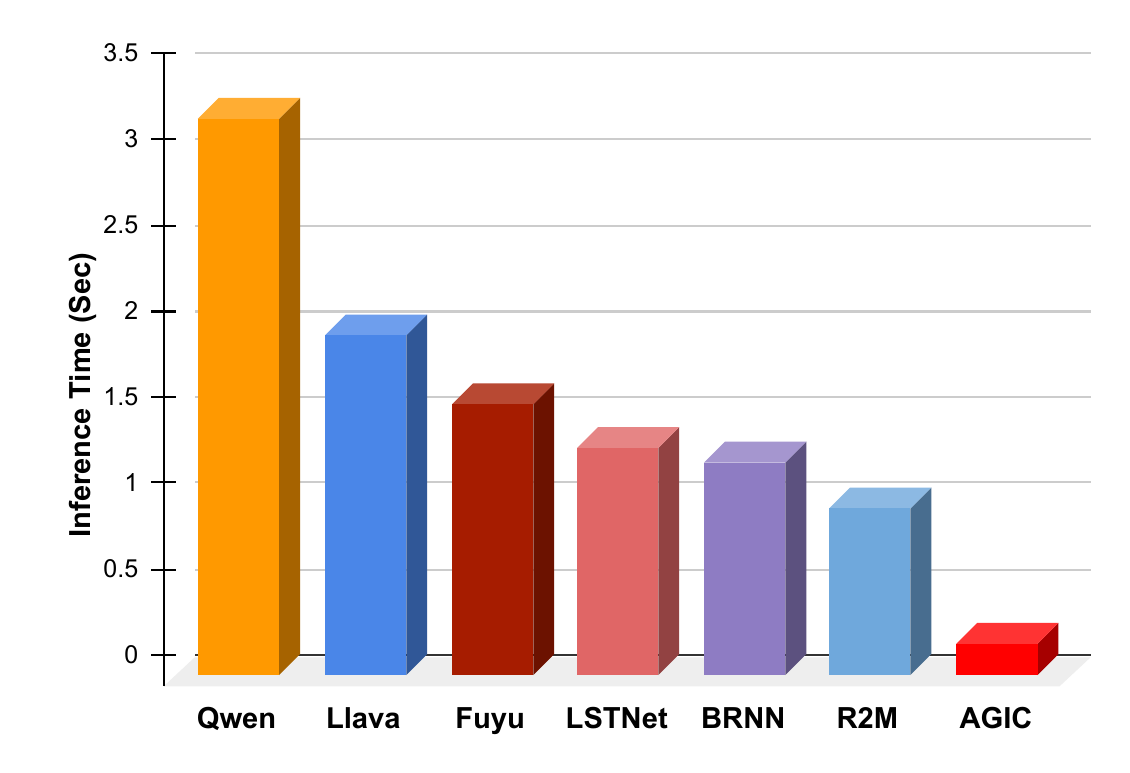}
    \caption{Inference time comparison.}
    \label{fig:inference_time}
    \vspace{-5mm}
\end{figure}

When comparing the inference time per sample with other state-of-the-art models, as shown in Figure~\ref{fig:inference_time}, our AGIC approach demonstrates significantly lower latency. Specifically, AGIC takes only 0.19 seconds per sample, whereas all zero-shot-based approaches require over 1.5 seconds. Additionally, we observe that the unsupervised method (R$^{2}$M) is faster than the supervised methods (BRNN and LSTNet). Overall, AGIC achieves the lowest inference time among all compared methods.

\section{Experimental Setup}
\label{sec:experimental_setup}
We conduct all our experiments on Amazon Web Services (\textit{AWS}) cloud server, Amazon elastic compute cloud (\textit{EC2}) instance. In the EC2 instance, we initiated an instance for accelerated Computing. The specifications are \textit{g6e.xlarge} instance, which provides 3rd generation AMD EPYC processors (\textit{AMD EPYC 7R13}), with a NVIDIA L40S Tensor Core GPU with 48 GB GPU memory, and 4x vCPU with 150 GiB memory. All the hyperparameter details are listed in Table~\ref{tab:hyperparams}. The model sources are detailed in Table~\ref{tab:model_sources}.

\begin{table}[h!]
    \centering
    \begin{tabular}{ll}
        \toprule
        \textbf{Hyperparameter} & \textbf{Value} \\
        \midrule
        \texttt{top\_p} & 0.9 \\
        \texttt{num\_return\_sequences} & 1 \\
        \texttt{num\_beams} & 5 \\
        \texttt{max\_new\_tokens} & 30 \\
        \texttt{top\_k} & 50 \\
        \texttt{early\_stoping} & True \\
        \texttt{do\_sample} & True \\
        \texttt{temperature} & 1.0\\
        \texttt{K (amplification factor)} & 1, 3, 5, 10 \\
        \bottomrule
    \end{tabular}
    \caption{Hyperparameters details.}
    \label{tab:hyperparams}
\end{table}

\begin{table}[h!]
    \centering
    \resizebox{\linewidth}{!}{%
    \begin{tabular}{ll}
        \toprule
        \textbf{Model} & \textbf{Source} \\
        \midrule
        blip2-opt-2.7b & \url{https://huggingface.co/Salesforce/blip2-opt-2.7b} \\
        Qwen2.5-VL-7B-Instruct & \url{https://huggingface.co/Qwen/Qwen2.5-VL-7B-Instruct} \\
        fuyu-8b & \url{https://huggingface.co/adept/fuyu-8b} \\
        llava-1.5-7b-hf & \url{https://huggingface.co/llava-hf/llava-1.5-7b-hf} \\
        \bottomrule
    \end{tabular}%
    }
    \caption{Source of Huggingface models.}
    \label{tab:model_sources}
    \vspace{-5mm}
\end{table}

\section{More details on ablation study}
\label{sec:ablation_more}

\subsection{Effect of various layers attention visualization}
We perform various experiments to find out the optimal layer to extract the attention visualizations and experimented with First, Middle, Last, Max, and Mean of attention layers. If there are $L$ attention layers then the first attention layer can be represented as $a^{0}$, and middle layer will be represented as $a^{M}$ where $M=L//2$, the last layer will be $a^{L-1}$, the layer with maximum attention can simply be determined by the following $a^{max} = \max_{l}a^{l}$, and finally the mean attention of all the layers can be computed as $a^{mean} = \frac{1}{L}\sum_{l=0}^{L-1}a^{l}$. We obtain the caption for the image in all the layers mentioned earlier. 

\begin{table*}[t]
    \centering\footnotesize
    \begin{tabular}{llccccccccc}
        \toprule
        \textbf{Dataset $\downarrow$} & \textbf{Model $\downarrow$} & \textbf{Layer $\downarrow$ Metric $\rightarrow$} & \textbf{B1} & \textbf{B2} & \textbf{B3} & \textbf{B4} & \textbf{METEOR} & \textbf{R-L} & \textbf{CIDEr} & \textbf{SPICE} \\
        \midrule
        \multirow{5}{*}{Flickr8k} 
        & \multirow{5}{*}{BLIP-2-OPT} & First & 0.64 & 0.47 & 0.34 & 0.24 & 0.24 & 0.43 & 0.70 & 0.19 \\
        & & Mid & 0.64 & 0.46 & 0.33 & 0.23 & 0.23 & 0.43 & 0.69 & 0.19 \\
        & & Last & 0.64 & 0.47 & 0.34 & 0.24 & 0.24 & 0.44 & 0.69 & 0.19 \\
        & & Max & 0.62 & 0.45 & 0.33 & 0.23 & 0.23 & 0.42 & 0.66 & 0.18 \\
        & & Mean & \textbf{0.66} & \textbf{0.49} & \textbf{0.35} & \textbf{0.25} & \textbf{0.24} & \textbf{0.44} & \textbf{0.73} & \textbf{0.19} \\
        \midrule
        \multirow{5}{*}{Flickr30k} 
        & \multirow{5}{*}{BLIP-2-OPT} & First & 0.62 & 0.45 & 0.32 & 0.22 & 0.21 & 0.38 & 0.56 & 0.15 \\
        & & Mid & 0.62 & 0.44 & 0.32 & 0.22 & 0.21 & 0.38 & 0.55 & 0.16 \\
        & & Last & 0.63 & 0.45 & 0.32 & 0.22 & 0.21 & 0.38 & 0.56 & 0.15 \\
        & & Max & 0.60 & 0.43 & 0.30 & 0.21 & 0.20 & 0.37 & 0.52 & 0.15 \\
        & & Mean & \textbf{0.65} & \textbf{0.46} & \textbf{0.33} & \textbf{0.23} & \textbf{0.23} & \textbf{0.39} & \textbf{0.60} & \textbf{0.16} \\
        \bottomrule
    \end{tabular}
    \caption{Evaluation metrics BLEU (1-4), METEOR, ROUGE-L, CIDEr, and SPICE on two datasets Flickr8k and Flickr30k, for the captions from different layer strategies.}
    \label{tab:diff_layers}
\end{table*}

We have evaluated our method AGIC on benchmark datasets: Flickr8k and Flickr30k, using standard image captioning metrics: BLEU-1 to BLEU-4 (B1–B4), METEOR, ROUGE-L, CIDEr, and SPICE. In Table~\ref{tab:diff_layers}, we report performance under different attention layers, decoding strategies.
On both datasets, the BLIP2, a 2.7 billion-parameter model, performs well. Still, its performance is significantly improved with the help of hybrid decoding strategies and the attention-guided amplification technique.

\subsection{Effect of Decoding Strategies}
On both the Flickr8k and Flickr30k datasets, combining the decoding strategies: 1) Top-k sampling, 2) Top-p (nucleus) sampling, and 3) Beam Search yielded substantial performance gains across all evaluation metrics compared to the base model. As shown in Table~\ref{tab:combined_ablation_studies}, the CIDEr score on Flickr8k improved from 0.19 (base) to 0.73 (All), and on Flickr30k from 0.43 to 0.60.

\begin{figure}[t]
    \centering
    \includegraphics[width=0.45 \textwidth]{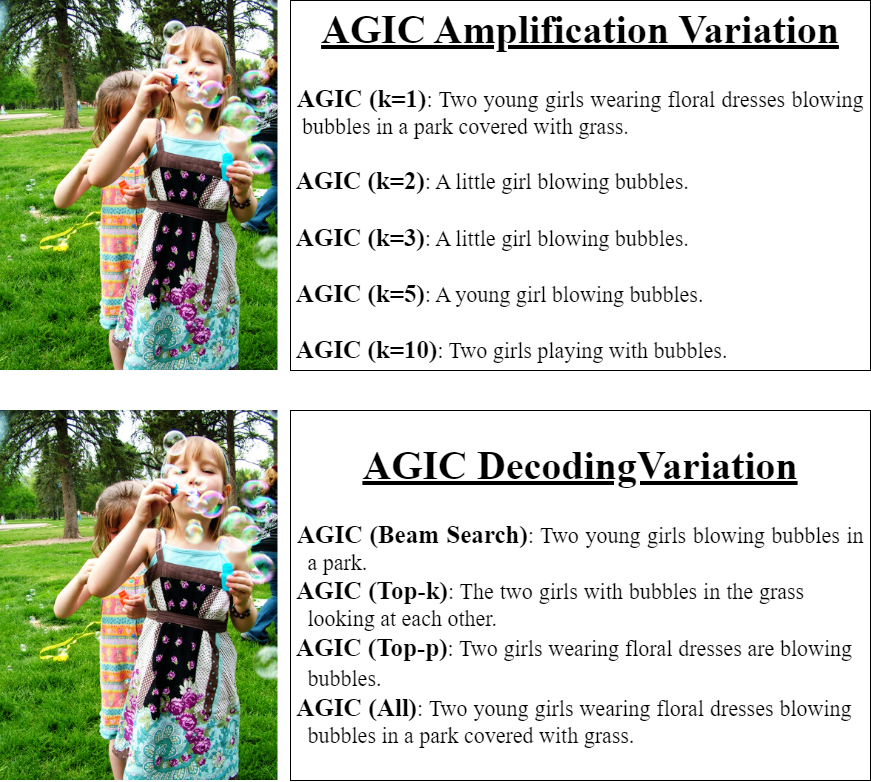} 
    \caption{Ablation on the effects of amplification strength and decoding strategy in the AGIC model for a sample image. The top frame illustrates the effect of varying the amplification factor $k$ (from $k{=}1$ to $k{=}10$) to guide the caption generation. The bottom frame compares different decoding strategies applied to the same attention-guided image: beam search, top-$k$ sampling, top-$p$ (nucleus) sampling, and ensemble of all decoding strategies.}
    \label{fig:AGIC_ablation}
\end{figure}

Among these, beam search played a pivotal role in enhancing caption quality, contributing the most significant improvement, particularly in BLEU and CIDEr scores. This can be attributed to its ability to maintain the most probable caption candidates across decoding steps, ensuring better fluency and grammatical correctness. However, beam search alone tends to produce less diverse captions. The inclusion of Top-p and Top-k sampling addresses this limitation by introducing diversity, which enhances recall-based metrics such as ROUGE-L and METEOR when used in conjunction with beam search. Figure~\ref{fig:AGIC_ablation} illustrates the qualitative impact of different decoding strategies on a sample image.

\subsection{Effect of Amplification Factor $k$}

The amplification factor $k$ controls how many times the resulting attention maps from the first pass are integrated into the image tensor for the amplification of important and relevant features in the image. As seen in the table \ref{tab:combined_ablation_studies}, $k=1$ yields the best results. This shows that using only the one prominent attention map effectively enhances key visual regions without introducing noise. As $k$ increases, performance begins to drop like on Flickr8k, CIDEr drops from 0.734 ($k=1$) to 0.565 ($k=5$). This is because including more attention weights tends to dilute the focus and overemphasize not only the relevant regions but also the less relevant or irrelevant regions, which leads to a decline in caption generation quality. The decrease in performance as the amplification factor increases is consistent across metrics, confirming that excessive amplification causes distraction in the model’s attention. An illustration of different values of amplification factor applied to an image is provided at the top frame of Figure~\ref{fig:AGIC_ablation}.

\subsection{Effect of caption length}
To explore how caption length correlates with evaluation metrics, we conduct a case study. Our observations show that longer captions generally yield slightly higher scores in CIDEr, METEOR, and SPICE, likely due to their ability to capture more contextual details. By grouping captions into length buckets (1–5, 6–10, 11–15), we find that:\\
a) Captions with 6–10 words strike a good balance between precision (BLEU) and semantic adequacy (SPICE, METEOR).\\
b) Very short captions (<5 words) often omit key details, resulting in lower completeness and SPICE scores.\\
c) Very long captions (>15 words) do not consistently improve performance and may introduce noise or redundancy, affecting correctness. \\
Overall, captions of moderate length (around 7–12 words) tend to achieve the best scores across both human and automatic evaluations.
\section{Algorithm}

\begin{algorithm}[H]
\caption{The AGIC Approach}
\label{alg:agic}

\textbf{Input:} IC model $\mathcal{M}$, image $I$, ref cap $C_{\text{ref}}$ \\
\textbf{Output:} Amplified cap $C_{\text{attn}}$, eval scores $E$

\vspace{0.5em}
1. Preprocess $I$ to obtain tensor $I_{\text{tensor}}$.

2. Pass $I_{\text{tensor}}$ through $\mathcal{M}$’s encoder to extract attention maps $A^{l,h}$ for all layers $l$ and heads $h$.

3. Compute patch attention scores $a_i^l$ from Eq \ref{mead_head_attn} and reshape into $a^l(x,y)$.

4. For each strat $\text{s} \in \{\text{first}, \text{mid}, \text{last}, \text{max}, \text{mean}\}$:

\begin{itemize}
  \item Determine $a^{\text{s}}(x,y)$ as acc to \ref{inbetween_passes}: \\
  \hspace*{1em} first: $a^0(x,y)$, \\
  \hspace*{1em} mid: $a^{M}(x,y)$, where $M=L//2$\\
  \hspace*{1em} last: $a^{L}(x,y)$, \\
  \hspace*{1em} max: $\max_{l} a^{l}(x,y)$, \\
  \hspace*{1em} mean: $\frac{1}{L} \sum_{l} a^{l}(x,y)$
  \item Amp: ${I_a}^{\text{s}}(x,y) = I_o(x,y) \cdot (1 + k \cdot a^{\text{s}}(x,y))$
  \item Gen caption $C_{\text{attn}}[\text{s}] = \mathcal{M}(\hat{I}^{\text{s}})$ acc to \ref{sec_pass}.
\end{itemize}

5. Evaluate $C_{\text{attn}}$ using BLEU-1 to BLEU-4, ROUGE-L, METEOR, CIDEr, and SPICE w.r.t.\ $C_{\text{ref}}$ acc to \ref{comparision}.

6. \textbf{Return:} $C_{\text{attn}}, E$
\end{algorithm}

\begin{figure*}[t]
    \centering
    \foreach \img/\gt/\agic/\err in {
        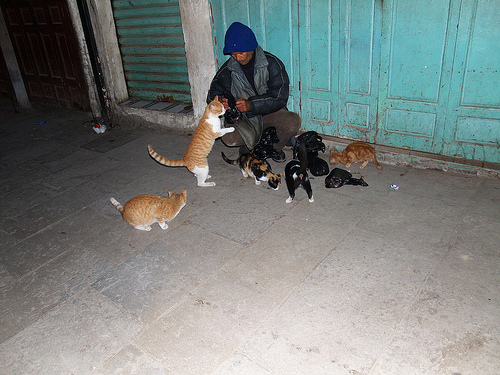/A man on the street surrounded by cats./A man petting cats and dogs on a sidewalk./\textcolor{red}{Hallucination. No dogs in the image.},
        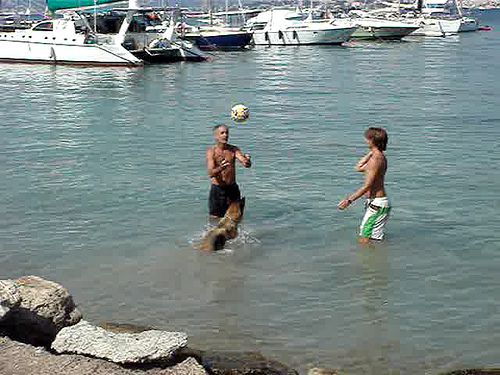/Two men and a dog are playing with a ball in front of the boats in the ocean./Two boys playing with a ball in the water with yachts in background./\textcolor{red}{Omission. The dog is not included.},
        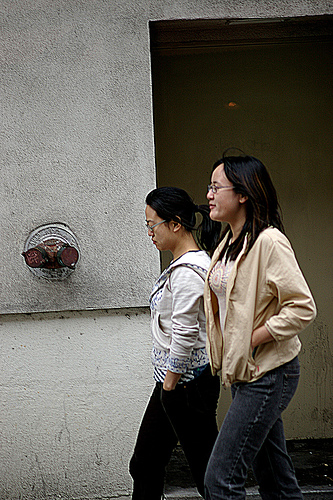/Two young women walk past a door in a white wall./Two women passing down the street beside a black door./\textcolor{red}{Irrelevant: No black door, only white wall.},
        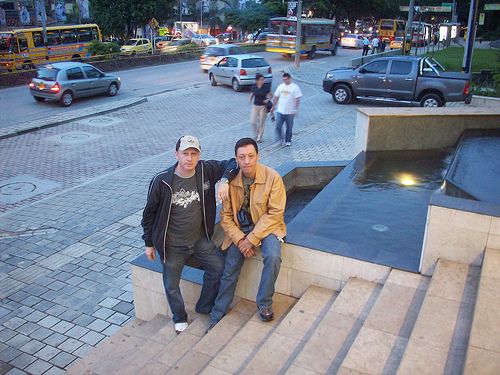/Two guys are posing next to each other on the steps of a building with traffic behind them./Two men sitting on steps./\textcolor{red}{Vague: The caption can be more descriptive besides sitting on steps.},
        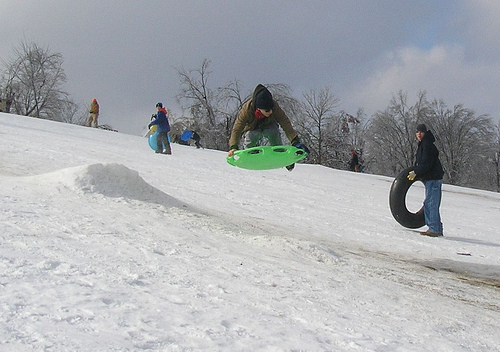/A child sleds over a mound of snow as others watch him./A person is watching a man snowboard down a hill./\textcolor{red}{Ambiguity: The caption mentioned it as just the person and a man, but couldn't identify the gender and age.},
        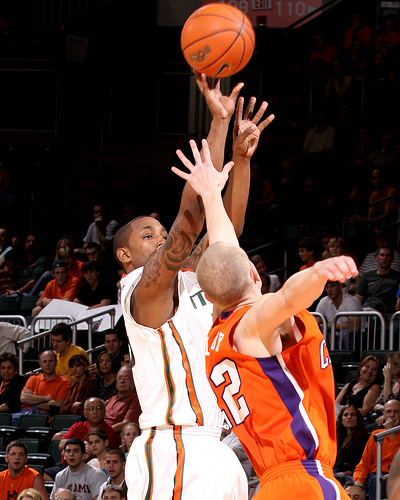/A basketball player shooting while another player is trying to block his shot./a basketball player trying to block the ball from going into the basket thrown by another player./\textcolor{green}{All Correct with no errors.}
    }{
        \begin{minipage}[t]{0.3\textwidth}
            \centering
            \includegraphics[height=3.8cm,valign=t]{\img}
            \vspace{0.3em}
            
            \captionsetup{justification=centering}
            \caption*{\scriptsize \textbf{\textit{GT}}: \gt\\ \textbf{\textit{AGIC}}: \agic\\ \textbf{\textit{Error}}: \err}
        \end{minipage}
        \hspace{0.03\textwidth}
        \ifnum\value{enumi}=2 \\[1.5em] \fi
    }
    \caption{Qualitative error analysis of image captions generated by the AGIC model compared to ground truth (GT) descriptions. Each example highlights a specific type of error: hallucination, omission, irrelevance, vagueness, and ambiguity. One example demonstrates a correct caption with no notable issues.}
    \label{fig:error_analysis_examples}
    \vspace{-5mm}
\end{figure*}

\section{More detail on Human Evaluation}
\label{sec:human_appendix}
To perform the human evaluation, we have given 50 samples to evaluators and instruct them to assess each of the parameter based on the following definition.\\
 \textbf{Correctness:} which assesses the accurate identification of all prominent objects; \\
 \textbf{Completeness:} which evaluates whether all relevant image content, including objects, attributes, and actions, is comprehensively covered; \\
 \textbf{Relevancy:} which measures the pertinence and salience of the information presented in relation to the image's primary content.

As detailed in Table~\ref{tab:icc_rating}, we compute the inter-rate reliability using Intra-class correlation coefficient as per ICC \cite{koo2016guideline} and observe that evaluator exhibits moderate to good agreement for both Flickr8k and Flickr30k data samples.

\begin{table}[ht]
\centering\footnotesize
\setlength{\tabcolsep}{0.8ex}
\begin{tabular}{lccc}
\toprule
 & \textbf{Relevancy} & \textbf{Correctness} & \textbf{Completion} \\
\midrule
\textbf{Flickr8k} & 0.80 & 0.86 & 0.75 \\
\textbf{Flickr30k} & 0.77 & 0.77 & 0.88 \\
\bottomrule
\end{tabular}
\caption{Inter rater reliability (ICC).}
\label{tab:icc_rating}
\end{table}

\section{More Details on Error Analysis}
\label{sec:ablation_error}

As mentioned in Subsection \ref{sec:human_error}, to evaluate the AGIC approach, we analyzed 50 image samples from the Flickr8k and Flickr30k datasets by considering four error categories: Hallucination, Omission, Irrelevance, and Ambiguity. 

\begin{enumerate}
\item \textbf{Hallucination:}
Hallucination refers to the inclusion of objects, actions, or details in the caption that are not actually present in the image. This typically occurs when models rely too heavily on prior knowledge or context rather than visual evidence, leading to inaccurate or fabricated descriptions.

\item \textbf{Omission:}  
Omission occurs when the caption fails to mention relevant visual elements that are clearly present in the image. While it may describe the main subject, it might overlook important background details or secondary objects, resulting in an incomplete representation of the scene.

\item \textbf{Irrelevance:}  
Irrelevance indicates a mismatch between the generated caption and the visual content of the image. It occurs when the caption includes information that is off-topic or not grounded in the image, reflecting poor alignment between vision and language.

\item \textbf{Ambiguity:}  
Ambiguity arises when the caption uses vague or generic terms to describe entities in the image—for example, using ``someone" or ``a person" without specifying characteristics like gender, role, or activity. This lack of clarity can make the caption less informative or confusing.
\end{enumerate}
Furthermore, illustrative examples for each of the above error categories are provided in Figure~\ref{fig:error_analysis_examples}.
\section{Mapping 1D Attention to 2D Spatial Layout}
\label{sec:map1dto2d}
The attention scores obtained during the \textit{Attention Weights Extraction} step (as mentioned in Section~\ref{first_pass}) are one-dimensional vectors, representing the mean attention across all heads in the model. Since these scores are 1D, we must transform them into a 2D spatial layout to apply amplification to the image. To achieve this, we utilize attention weights associated with the [CLS] token, which serves as a global aggregator of information across the image. The resulting 1D attention vector is then reshaped into a 2D attention map corresponding to the spatial dimensions of the preprocessed image (typically forming a square grid).
.

\section{Prompt for zero-shot captioning}
We perform image captioning using a zero-shot prompting strategy with vision-language models (VLLMs). The prompt template shown in Figure~\ref{fig:zero_shot_prompt} is used for zero-shot caption generation with models such as LLaVA, Qwen, and Fuyu.

\begin{figure}[htb]
\centering
\begin{tabular}{@{}p{\linewidth}@{}}
\toprule
\textbf{Zero-shot Prompt}\\
\midrule
\texttt{\{"role": "user", "content": [} \\
\texttt{\{"type": "image", "image": <image>\},} \\
\texttt{\{"type": "text", "text": "Generate a COCO-style caption focused on the main objects and their interactions. Avoid names; keep it concise and grammatically correct."\}} \\
\texttt{]\}} \\
\bottomrule
\end{tabular}
\caption{Zero-shot prompt used for caption generation.}
\label{fig:zero_shot_prompt}
\end{figure}

\end{document}